\documentclass[11pt,twocolumn]{article}

\usepackage[utf8]{inputenc}
\usepackage[T1]{fontenc}
\usepackage{times}
\usepackage[margin=0.75in,columnsep=0.25in]{geometry}

\usepackage{amsmath,amssymb,amsfonts}

\usepackage{booktabs}
\usepackage{multirow}
\usepackage{array}
\usepackage{tabularx}

\usepackage{graphicx}
\usepackage{tikz}
\usepackage{pgfplots}
\pgfplotsset{compat=1.18}
\usetikzlibrary{arrows.meta,positioning,shapes.geometric,fit,calc,backgrounds,decorations.pathreplacing}
\usepackage{subcaption}
\usepackage{caption}
\usepackage[dvipsnames]{xcolor}

\usepackage[ruled,vlined,linesnumbered]{algorithm2e}

\usepackage{enumitem}
\usepackage{hyperref}
\hypersetup{colorlinks=true,linkcolor=MidnightBlue,citecolor=MidnightBlue,urlcolor=MidnightBlue}
\usepackage{cleveref}
\usepackage{url}
\usepackage{microtype}
\usepackage{float}
\usepackage{stfloats}
\usepackage[section]{placeins}

\definecolor{stateblue}{HTML}{2E5090}
\definecolor{agentgreen}{HTML}{2D8F4E}
\definecolor{toolorange}{HTML}{D4761C}
\definecolor{judgeviolet}{HTML}{7B2D8E}
\definecolor{userteal}{HTML}{1A8A8A}
\definecolor{lightgray}{HTML}{F0F0F0}

\usepackage{titlesec}
\titlespacing*{\section}{0pt}{2.0ex plus 0.5ex minus .2ex}{1.0ex plus .2ex}
\titlespacing*{\subsection}{0pt}{1.5ex plus 0.3ex minus .2ex}{0.8ex plus .2ex}
\titlespacing*{\subsubsection}{0pt}{1.0ex plus 0.2ex minus .1ex}{0.5ex plus .1ex}

\setlist{nosep,leftmargin=*}

\newcommand{\sys}{\textsc{StateGen}}

\newcommand{\eg}{\textit{e.g.}}

\begin{document}


\title{\Large \textbf{State-Grounded Multi-Agent Synthetic Data Generation\\for Tool-Augmented LLMs}}

\author{%
Rahul Khedar$^*$, Eshita$^*$, Sneha Teja Sree Reddy Thondapu$^*$, Mayank Malhotra$^*$\\
Arup Das$^\dagger$, Jitesh Chandra$^\dagger$, Yun-Shiuan Chuang$^\dagger$, Chaitanya Kulkarni$^\dagger$, Arun Menon$^\dagger$, Linsey Pang, \\Avinash Karn, Mouli V, Prakhar Mehrotra\\[8pt]
\textit{PayPal AI}\\[4pt]
}
\date{}
\maketitle

\begin{abstract}
Training tool-augmented LLM agents requires large corpora of multi-turn, tool-grounded conversational data that is expensive to annotate, privacy-constrained in production settings, and largely absent from public datasets. We present \sys{}, a synthetic data generation platform that produces scored, reasoning-trace-rich training conversations by orchestrating a four-role LLM loop: a \emph{persona-conditioned user simulator}, an \emph{agent under test}, a \emph{state-grounded tool simulator}, and a \emph{multi-axis LLM judge}. The key architectural contribution is an \textbf{authoritative state manager} that maintains a structured world-state object across turns, enforcing a backend-is-truth invariant that eliminates the dominant class of tool-call hallucinations by construction. \sys{} extends naturally to \textbf{hierarchical multi-agent} settings by declaring sub-agents as tools, all sharing a single state object. We report results on 64,698 evaluated conversations across three production corpora: tool-call hallucination scores reach 9.66/10, the system supports persona-driven variation via a 23-dimensional trait vector, and a cleanly separated train and golden evaluation set split confirms the data is not memorization bait (per-criterion gap analysis). Comparison with eight external systems shows that no single publicly available platform combines multi-turn generation, state-grounded tool simulation, hierarchical multi-agent support, and built-in judge scoring.
\end{abstract}


\section{Introduction}
\label{sec:intro}

The rapid deployment of LLM-based agents that interact with external tools like APIs, databases, CRM systems, payment processors has created an acute data bottleneck. Fine-tuning a strong function-calling model for a single business domain requires thousands of multi-turn, tool-grounded conversations covering the long tail of user intents, edge cases, and adversarial inputs. Human annotators produce 10--40 examples per hour, the majority of which fail quality review, making a 10K-sample dataset a multi-month project. Meanwhile, real conversation logs from production systems contain PII, payment credentials, and sensitive metadata that make them unsuitable as training data without extensive redaction.

Public datasets offer little help for agentic training. ShareGPT, Dolly, and similar corpora contain almost no tool-use trajectories. Benchmarks like $\tau$-bench \cite{yao2024taubench} and AgentBench \cite{liu2023agentbench} provide hundreds of trajectories for evaluation but not the tens of thousands needed for fine-tuning. The gap between evaluation-scale and training-scale agentic data remains largely unaddressed.

We present \sys{}, a synthetic data generation platform that addresses this gap through four design principles:

\begin{enumerate}[label=\textbf{(\arabic*)}]
    \item \textbf{State-grounded tool simulation.} An authoritative state manager maintains a structured world-state object that every simulated tool response must be consistent with. This invariant backend is the source of truth that eliminates the class of hallucinations where a tool ``discovers'' records that do not exist (\Cref{sec:state}).
    \item \textbf{Multi-role generation loop.} Four LLM roles (user simulator, agent under test, tool simulator, judge) run in a coordinated loop, each with independent model assignment and temperature, producing conversations that are simultaneously realistic, tool-grounded, and scored (\Cref{sec:architecture}).
    \item \textbf{Hierarchical multi-agent support.} Sub-agents are declared as tools, sharing a single state object through the state manager. This enables training data generation for orchestrator--sub-agent systems without hand-written simulation code (\Cref{sec:multiagent}).
    \item \textbf{Persona-conditioned user simulation.} A three-tier persona vector (6 categorical demographics, 12 continuous behavioral traits, 5 scenario-reactive emotional states) drives measurable variation in generated conversations (\Cref{sec:persona}).
\end{enumerate}

\paragraph{Contributions.} (i) We formalize a state-grounded tool simulation paradigm where the world-state object is the single source of truth, achieving 9.66/10 on tool-call hallucination across 49K+ samples. (ii) We introduce a hierarchical multi-agent generation architecture where sub-agents are tools sharing authoritative state. (iii) We propose a three-tier persona vector with 23 dimensions and demonstrate a 15.8 percentage-point spread in goal-achievement across personas. (iv) We present an 8-axis LLM judge with per-project extensibility and show that hallucination axes are empirically independent from goal-achievement ($\rho \approx 0.21$). (v) We compare against eight external systems and identify the capability gap that motivates this work.


\section{Related Work}
\label{sec:related}

\paragraph{Synthetic data for LLMs.}
Self-Instruct \cite{wang2023selfinstruct} and Alpaca \cite{taori2023alpaca} showed that LLM-generated instruction data can improve model performance, but these are single-turn and lack tool grounding. WizardLM \cite{xu2023wizardlm} introduced complexity-driven evolution but remained in the text-only domain. Phi-1 \cite{gunasekar2023textbooks} demonstrated the value of high-quality synthetic data at scale for code generation. None of these address multi-turn, tool-grounded conversations.

\paragraph{Tool-augmented LLM agents.}
Toolformer \cite{schick2023toolformer} taught models when and how to call APIs. ReAct \cite{yao2023react} interleaved reasoning and action. Function-calling conventions from OpenAI and open-source equivalents in vLLM established the dominant API pattern. These works focus on inference-time tool use and our contribution is on the \emph{training data generation} side.

\paragraph{LLM-based evaluation and judging.}
MT-Bench \cite{zheng2023judging} introduced pairwise LLM judging for chatbot quality. $\tau$-bench \cite{yao2024taubench} proposed task-level evaluation for tool-agent-user interaction with hand-curated database states. Our judge operates on eight independent axes with per-project extensibility, and it runs \emph{inside} the generation loop rather than as a post-hoc evaluation.

\paragraph{Multi-agent systems.}
AutoGen \cite{wu2023autogen}, CrewAI, and LangGraph provide runtime orchestration for multi-agent LLM applications. Meta's Matrix \cite{mei2024matrix} is the closest work on multi-turn multi-agent \emph{data generation}, but it remains a research prototype without an explicit shared-state object. AgentBench \cite{liu2023agentbench} curates evaluation trajectories across eight environments but does not generate training data.

\paragraph{Reasoning distillation.}
DeepSeek-R1 \cite{deepseek2025r1} demonstrated that chain-of-thought distillation from a large reasoning model into smaller students yields strong gains on math and code benchmarks. Our platform captures reasoning traces from any model with an OpenAI-compatible endpoint and exports them in fine-tune-ready JSONL, enabling the same distillation pattern for agentic, tool-grounded domains.

\paragraph{Positioning.}
The distinguishing feature of \sys{} is the \emph{combination}: multi-turn generation, state-grounded tool simulation, hierarchical multi-agent support, persona conditioning, and built-in multi-axis judging. Individual capabilities exist in isolation across the systems cited above and no current single platform ships all five.


\section{System Architecture}
\label{sec:architecture}

\begin{figure*}[t]
\centering
\begin{tikzpicture}[
    node distance=1.6cm and 2.2cm,
    role/.style={draw, rounded corners=6pt, minimum width=2.8cm, minimum height=1.0cm, font=\small\bfseries, text=white, align=center},
    state/.style={draw=stateblue, fill=stateblue!10, rounded corners=4pt, minimum width=2.4cm, minimum height=0.8cm, font=\small\bfseries, align=center},
    arrow/.style={-{Stealth[length=6pt]}, thick},
    data/.style={draw=gray!60, dashed, rounded corners=3pt, minimum width=2.0cm, minimum height=0.6cm, font=\footnotesize, align=center},
]
\node[role, fill=userteal] (user) {User\\Simulator};
\node[role, fill=agentgreen, right=of user] (agent) {Agent Under\\Test};
\node[role, fill=toolorange, right=of agent] (tool) {Tool\\Simulator};
\node[role, fill=judgeviolet, below=2.5cm of agent] (judge) {LLM Judge\\(8 Axes)};
\node[state, below=1.0cm of tool] (statemgr) {State\\Manager};
\node[draw=stateblue, fill=stateblue!5, rounded corners=3pt, minimum width=1.8cm, minimum height=0.6cm, font=\footnotesize, align=center, right=1.6cm of statemgr] (worldstate) {World\\State $\mathcal{S}_t$};
\node[data, above=0.8cm of user] (persona) {Persona $\mathbf{p}$};
\node[data, above=0.8cm of agent] (playbook) {Playbook};
\node[data, above=0.8cm of tool] (scenario) {Scenario\\$(\mathcal{F}_u, \mathcal{F}_s, g, \mathcal{S}^*)$};
\node[data, below=0.8cm of judge] (output) {Scored JSONL\\+ Reasoning Traces};
\draw[arrow, userteal] (user) -- node[above, font=\scriptsize] {query $q_t$} (agent);
\draw[arrow, agentgreen] (agent) -- node[above, font=\scriptsize] {tool call $a_t$} (tool);
\draw[arrow, toolorange] (tool) -- node[below, font=\scriptsize, yshift=-2pt] {response $r_t$} (agent);
\draw[arrow, stateblue] (tool) -- node[right, font=\scriptsize] {$(a_t, r_t)$} (statemgr);
\draw[arrow, stateblue] (statemgr) -- node[above, font=\scriptsize] {update} (worldstate);
\draw[arrow, stateblue, dashed] (worldstate) |- node[near start, right, font=\scriptsize] {$\mathcal{S}_t$} ([yshift=0.3cm]tool.south east);
\draw[arrow, gray] (persona) -- (user);
\draw[arrow, gray] (playbook) -- (agent);
\draw[arrow, gray] (scenario) -- (tool);
\draw[arrow, gray, dashed] (scenario.south) -- ++(0,-0.4) -| (statemgr.north);
\draw[arrow, judgeviolet, dashed] ([xshift=-0.5cm]agent.south) |- node[near start, right, font=\scriptsize] {transcript} (judge);
\draw[arrow, judgeviolet] (judge) -- (output);
\draw[arrow, userteal, dashed] (agent.south west) -- ++(0,-0.3) -| node[near start, below, font=\scriptsize] {response} (user.south);
\end{tikzpicture}
\caption{The \sys{} four-role architecture. The \textbf{state manager} (center right) maintains a world-state object $\mathcal{S}_t$ that the tool simulator must be consistent with. After the conversation completes, the judge scores the full transcript on 8 independent axes.}
\label{fig:architecture}
\end{figure*}
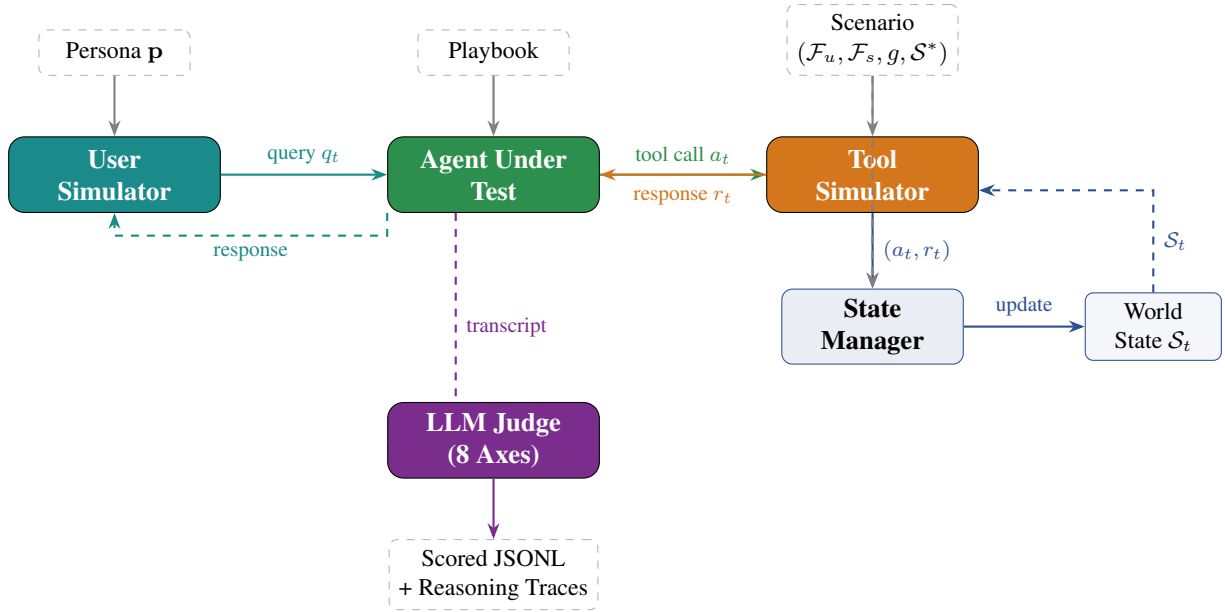

\sys{} converts a scenario specification, a persona vector, and a playbook into a scored, reasoning-trace-rich conversation. The core is a four-role conversational loop: user simulator, agent under test, tool simulator, and judge, supported by a fifth LLM, the \emph{authoritative state manager}, which maintains world-state consistency but does not participate in the conversation itself. \Cref{fig:architecture} illustrates the architecture.

\subsection{Input Specification}
\label{sec:inputs}

Each generation run requires three inputs:

\textbf{Scenario.} A structured YAML document specifying: user facts $\mathcal{F}_u$ (what the customer knows), system facts $\mathcal{F}_s$ (ground-truth backend state including balances, records, KYC status), a user goal $g$ (success condition from the user's perspective), and a final state $\mathcal{S}^*$ (expected backend changes and agent actions). The system facts $\mathcal{F}_s$ initialize the state manager's world-state object $\mathcal{S}_0$.

\textbf{Persona vector.} A 23-dimensional vector $\mathbf{p}$ comprising 6 categorical demographics, 12 continuous behavioral traits, and 5 continuous emotional states (\Cref{sec:persona}).

\textbf{Playbook.} An orchestrator system prompt (Jinja2 template), tool definitions with JSON schemas, per-tool capability templates that instruct the tool simulator, and an optional guardrail library.

\subsection{The Four Roles}
\label{sec:roles}

\paragraph{User simulator.} Reads the scenario and persona vector $\mathbf{p}$, producing a realistic turn-by-turn query stream. Persona traits drive vocabulary, impatience, assertiveness, and when to push back.

\paragraph{Agent under test.} The production-shape model that will eventually be fine-tuned. \sys{} supports both OpenAI-style function-calling and ReAct-style \cite{yao2023react} agents. The rest of the loop is architecture-agnostic.

\paragraph{Tool simulator.} For every tool call the agent issues, a second LLM receives the tool's Jinja2 capability template and the \emph{current world state} $\mathcal{S}_t$. It must respond consistently with $\mathcal{S}_t$, this is the state-grounding invariant (\Cref{sec:state}).

\paragraph{LLM judge.} After each conversation, the judge scores the transcript on eight independent criteria on a 0--10 scale with per-criterion rationale (\Cref{sec:judge}).

\subsection{Per-Role Model Routing}
\label{sec:routing}

Each role is bound to a model independently through a configuration registry. This separation serves two purposes: (i) the judge should not be the same model family as the agent to avoid score drift toward the model's own preferences; (ii) using different cost tiers per role makes the economics of 50K+ sample runs tractable.

\begin{table}[t]
\centering
\small
\begin{tabular}{@{}lcc@{}}
\toprule
\textbf{Role} & \textbf{Temp.} & \textbf{Rationale} \\
\midrule
User simulator    & 0.7 & Natural linguistic variation \\
Agent under test  & 0.7 & Match production inference \\
Tool simulator    & 0.3 & State coherence priority \\
State manager     & 0.0 & Deterministic state updates \\
LLM judge         & 0.2 & Slight randomness for tie-breaking \\
\bottomrule
\end{tabular}
\caption{Default per-role temperatures. The state manager runs at $\tau{=}0$ to prevent state drift.}
\label{tab:temperatures}
\end{table}


\section{State-Grounded Tool Simulation}
\label{sec:state}

The state manager is the architectural element that distinguishes \sys{} from prior synthetic data systems. We formalize its operation and the invariant it enforces.

\subsection{Formal Specification}

Let $\mathcal{S}_0 = \mathcal{F}_s$ be the initial world state loaded from the scenario's system facts. At each tool invocation step $t$, the agent issues a tool call $a_t = (\texttt{name}, \texttt{args})$. The tool simulator produces a response $r_t$ conditioned on both the capability template $\mathcal{T}_{\texttt{name}}$ and the current state:
\begin{equation}
    r_t \sim \text{LLM}_{\text{tool}}(\mathcal{T}_{\texttt{name}}, \mathcal{S}_{t-1}, a_t)
    \label{eq:tool_sim}
\end{equation}

The state manager then updates the world state:
\begin{equation}
    \mathcal{S}_t = \text{LLM}_{\text{state}}(\mathcal{S}_{t-1}, a_t, r_t)
    \label{eq:state_update}
\end{equation}

\subsection{The Backend-is-Truth Invariant}

The critical design constraint is:

\begin{quote}
\textit{A fact enters $\mathcal{S}_t$ if and only if it either (a) was in $\mathcal{S}_0 = \mathcal{F}_s$, or (b) was legally created by a tool call $a_j$ ($j \leq t$) with write authority.}
\end{quote}

This invariant is enforced by the state manager's system prompt, which explicitly instructs: ``If the state says a transaction exists, you must find it; if the state says it does not exist, you must return an error consistent with that.'' The tool simulator receives $\mathcal{S}_{t-1}$ and is similarly instructed to respect it as the sole source of truth.

\subsection{Why This Matters}

Without state grounding, a tool simulator LLM can hallucinate records, transactions, or system responses that have no basis in the scenario. This is the dominant failure mode in prompt-only tool simulation: the tool LLM generates a plausible-looking response that references a nonexistent account or fabricates a transaction amount. Since each turn builds on previous tool responses, a single hallucinated fact propagates through the entire conversation.

The state manager breaks this chain. By maintaining an explicit, structured world-state object that every tool response must be consistent with, the system prevents the fabricated facts from accumulating across turns. Empirically, this achieves a tool-call hallucination score of 9.66/10 across 49,331 evaluated conversations (\Cref{sec:results}).


\section{Hierarchical Multi-Agent Generation}
\label{sec:multiagent}

\sys{}'s most distinctive capability is generating training data for hierarchical multi-agent systems, not just single agents. \Cref{fig:multiagent} illustrates the pattern.

\begin{minipage}{\columnwidth}
\centering
\begin{tikzpicture}[
    node distance=0.5cm and 0.9cm,
    box/.style={draw, rounded corners=4pt, minimum width=1.7cm, minimum height=0.6cm, font=\footnotesize, align=center},
    orch/.style={box, fill=agentgreen!15, draw=agentgreen, font=\footnotesize\bfseries},
    sub/.style={box, fill=toolorange!12, draw=toolorange},
    st/.style={box, fill=stateblue!12, draw=stateblue, minimum width=2.2cm},
    arrow/.style={-{Stealth[length=4pt]}, thick},
]
\node[orch] (orch) {Orchestrator};
\node[sub, below left=0.9cm and 0.9cm of orch] (disputes) {\scriptsize Disputes\\[-1pt]\scriptsize Sub-Agent};
\node[sub, below=0.9cm of orch] (finance) {\scriptsize Finance\\[-1pt]\scriptsize Sub-Agent};
\node[sub, below right=0.9cm and 0.9cm of orch] (integ) {\scriptsize Integration\\[-1pt]\scriptsize Sub-Agent};
\node[st, below=0.7cm of finance] (state) {\scriptsize Shared State $\mathcal{S}$};
\draw[arrow, agentgreen] (orch) -- node[left, font=\tiny, pos=0.4] {\texttt{call\_*}} (disputes);
\draw[arrow, agentgreen] (orch) -- node[right, font=\tiny, pos=0.4] {\texttt{call\_*}} (finance);
\draw[arrow, agentgreen] (orch) -- node[right, font=\tiny, pos=0.4] {\texttt{call\_*}} (integ);
\draw[arrow, stateblue, densely dashed] (disputes) -- (state);
\draw[arrow, stateblue, densely dashed] (finance) -- (state);
\draw[arrow, stateblue, densely dashed] (integ) -- (state);
\node[font=\tiny, gray, above right=0.1cm and 0.2cm of orch] {Agent under test};
\node[font=\tiny, gray, below=0.12cm of state] {read \& write via state manager};
\draw[decorate, decoration={brace, amplitude=4pt, mirror}, thick, toolorange!70]
    ([xshift=-0.2cm]disputes.south west) -- ([xshift=0.2cm]integ.south east)
    node[midway, below=6pt, font=\tiny, text=toolorange!80!black]
    {tool simulator + capability templates};
\end{tikzpicture}
\captionof{figure}{Hierarchical multi-agent generation. Sub-agents are declared as tools sharing a single authoritative state object $\mathcal{S}$, ensuring cross-agent consistency.}
\label{fig:multiagent}
\end{minipage}

\subsection{Sub-Agents as Tools}

The pattern is sub-agents are declared as tools in the orchestrator's tool definitions. The orchestrator calls \texttt{call\_disputes\_agent(subquery=...)} the same way it would call any other tool. The tool simulator, given the appropriate capability template, responds as if it were the sub-agent's backend.

\begin{algorithm}[t]
\small
\SetAlgoLined
\SetKwInOut{Input}{Input}
\SetKwInOut{Output}{Output}
\Input{Scenario $(\mathcal{F}_u, \mathcal{F}_s, g, \mathcal{S}^*)$, Persona $\mathbf{p}$, Playbook $\Pi$}
\Output{Scored conversation $\mathcal{C}$, judge vector $\mathbf{j} \in \mathbb{R}^8$}
$\mathcal{S} \leftarrow \mathcal{F}_s$\;
$\mathcal{C} \leftarrow []$\;
\For{$t = 1$ \KwTo $T_{\max}$}{
    $q_t \leftarrow \text{LLM}_{\text{user}}(\mathbf{p}, \mathcal{F}_u, g, \mathcal{C})$ \tcp*{User sim}
    $\mathcal{C}.\text{append}(\texttt{user}, q_t)$\;
    \For{$s = 1$ \KwTo $S_{\max}$}{
        $a_t^s \leftarrow \text{LLM}_{\text{agent}}(\Pi, \mathcal{C})$ \tcp*{Agent}
        \If{$a_t^s$ is terminal action}{
            $\mathcal{C}.\text{append}(\texttt{assistant}, a_t^s)$\;
            \textbf{break outer loop}\;
        }
        $r_t^s \leftarrow \text{LLM}_{\text{tool}}(\mathcal{T}_{a_t^s.\texttt{name}}, \mathcal{S}, a_t^s)$ \tcp*{Tool}
        $\mathcal{S} \leftarrow \text{LLM}_{\text{state}}(\mathcal{S}, a_t^s, r_t^s)$ \tcp*{State}
        $\mathcal{C}.\text{append}(\texttt{tool}, r_t^s)$\;
    }
}
$\mathbf{j} \leftarrow \text{LLM}_{\text{judge}}(\mathcal{C}, \mathcal{S}^*)$ \tcp*{8-axis scoring}
\Return{$\mathcal{C}, \mathbf{j}$}
\caption{The \sys{} generation loop. $T_{\max}=10$ (turns), $S_{\max}=5$ (tool calls per turn). Sub-agent calls in multi-agent mode use the same tool-call pathway, with capability templates replacing sub-agent code.}
\label{alg:generation}
\end{algorithm}

\subsection{Shared State Across Sub-Agents}

The hidden invariant that makes hierarchical generation work is that \textbf{every sub-agent shares one authoritative state object}. When the orchestrator routes to a finance sub-agent, then to a disputes sub-agent, both read and write the same $\mathcal{S}_t$. Without shared state, the two sub-agents can return mutually incompatible facts at turn 5--10, producing training data that teaches the agent to tolerate contradictions.

\subsection{Zero-Code Sub-Agent Authoring}

Sub-agent capability templates can be generated from natural-language specifications via a knowledge engine module. An engineer describes the sub-agent's scope, capabilities, and constraints in structured YAML; the system produces Jinja2 capability templates, tool definitions, and configuration files. This reduces the authoring cost from Python implementation to specification review.


\section{Three-Tier Persona Model}
\label{sec:persona}

Persona is where \sys{} gets its variation. The persona vector $\mathbf{p}$ has three tiers: six categorical \emph{demographic} attributes, twelve continuous \emph{behavioral} traits in $[0,1]$, and five continuous \emph{emotional-state} traits that are scenario-reactive. Orthogonal to persona, a 4-level query-complexity overlay controls how the user simulator phrases each turn.

\subsection{Tier 1: Demographics (Categorical)}

Six attributes namely jurisdiction, age bracket, channel, device type, language proficiency, time availability, each drawn independently from a categorical distribution:
\begin{equation}
    a_k \sim \text{Categorical}(\mathbf{P}_k), \quad P_{k,j} = p_{k,j} / \textstyle\sum_m p_{k,m}
\end{equation}
Each category carries a prompt-guidance string injected into the user simulator's system prompt (\eg{}, jurisdiction=IN adds ``Uses INR, familiar with UPI'').

\subsection{Tier 2: Behavioral Traits (Continuous)}

Twelve traits namely cost sensitivity, patience, assertiveness, verbosity, politeness, domain knowledge, risk tolerance, compliance tendency, platform trust, digital literacy, slang usage, emoji usage are sampled around a hand-curated base vector $\mathbf{t}^{\text{base}}$ from a named profile:
\begin{equation}
    t_i = \text{clip}\big(t_i^{\text{base}} + \varepsilon_i,\; 0,\; 1\big), \quad \varepsilon_i \sim \mathcal{N}(0, \sigma^2), \quad \sigma = 0.08
    \label{eq:tier2}
\end{equation}

The choice of $\sigma = 0.08$ ensures 95\% of samples stay within $\pm 0.157$ of the base value, tight enough to remain in-persona but wide enough to avoid duplicate-looking samples. Continuous values are \textbf{bucketed} into \{low, medium, high\} using fixed breakpoints:
\begin{equation}
    \text{bucket}(v) = \begin{cases} \text{low} & v < 0.35 \\ \text{medium} & 0.35 \leq v < 0.70 \\ \text{high} & v \geq 0.70 \end{cases}
    \label{eq:buckets}
\end{equation}

Each bucket maps to a hand-written prompt-guidance string per trait.

\subsection{Tier 3: Emotional State (Scenario-Reactive)}

Five states: frustration, anxiety, trust, confidence, stress are initialized from a profile-specific range and then shifted by scenario-dependent deltas:
\begin{equation}
    e_i = \text{clip}\big(\mathcal{U}(l_i, h_i) + \Delta_i(\text{scenario}),\; 0,\; 1\big)
    \label{eq:tier3}
\end{equation}

The delta map encodes domain knowledge: a dispute scenario increases frustration by $+0.25$ and decreases trust by $-0.20$, making the same persona behave differently across scenarios.

\subsection{Trait Correlations}

Correlations are introduced through two mechanisms: (i) \emph{profile-encoded}, where correlated traits are co-set in the base vector by design; (ii) \emph{rule-described}, where post-sampling rules emit natural-language correlation statements into the prompt when threshold pairs are crossed (\eg{}, ``High digital literacy correlates with high domain knowledge'' fires only when both are $\geq 0.70$). Seven such rules are shipped. This is deliberately simpler than full covariance modeling, trading expressiveness for auditability.

\subsection{Query-Complexity Overlay}

Independent of persona, each sample draws a complexity tier: \textit{simple} (3--8 words, direct intent), \textit{medium} (8--15 words), \textit{complex} (15--30 words, backstory), or \textit{vague} (3--15 words, intent guessable but never named). The \textit{vague} tier is critical for agent training: it forces multi-turn clarification before tool selection, matching the distribution of real user traffic.


\section{Multi-Axis LLM Judge}
\label{sec:judge}

\sys{} does not produce a single quality score. It produces a vector of eight independent scores $\mathbf{j} \in \{1,\ldots,10\}^8$, plus rationale, enabling downstream filtering by axis.

\subsection{The Eight Default Criteria}

\Cref{tab:criteria} lists the eight criteria shipped with every project. Each produces an integer score in $\{1,\ldots,10\}$ with a per-criterion rationale.

\begin{table}[t]
\centering
\small
\begin{tabular}{@{}lp{4.8cm}@{}}
\toprule
\textbf{Criterion} & \textbf{What it measures} \\
\midrule
Goal achievement      & Did the agent accomplish the user's goal? \\
Tool usage            & Right tool, right arguments, right sequence? \\
Tool-call halluc.     & Did the agent invent tool calls, parameters, or results? \\
Reasoning quality     & Is the chain-of-thought coherent, on-policy, necessary? \\
Reasoning halluc.     & Did the agent reason about facts not in scope? \\
Communication quality & Clarity, tone, length, format for the persona \\
Consistency           & Does the agent contradict itself across turns? \\
Error handling        & Gracefulness under tool failures, pushback, or ambiguity \\
\bottomrule
\end{tabular}
\caption{The eight default judge criteria. Projects can add domain-specific axes (\eg{}, booking-flow compliance for CRM agents).}
\label{tab:criteria}
\end{table}

\subsection{Why Eight, Not One}

A single aggregate score conflates distinct failure modes. A conversation where the agent solves the problem but hallucinates a tool along the way is a failure for production deployment, even if the final message reads well. Splitting axes lets data consumers filter for their specific need: safety training, politeness, tool reliability. Empirically, the two hallucination axes form their own cluster ($\rho = 0.66$ pairwise) but correlate only weakly with goal achievement ($\rho \approx 0.21$), confirming hallucination is its own failure mode (\Cref{sec:results}).

\subsection{Per-Project Extensibility}

Each project can add, remove, or re-weight criteria. In our CRM case study, two domain-specific axes were added: \emph{booking-flow compliance} (mean 5.73 on training, 4.93 on golden) and \emph{channel-safety compliance} (mean 9.37 on training, 9.01 on golden). The former captures workflow-specific failure modes; the latter captures safety invariants. Different business concerns get different numbers.

\subsection{Scoring Protocol}

The judge LLM receives the full conversation transcript and the expected final state $\mathcal{S}^*$. It returns per-criterion integer scores $c_1, \ldots, c_8 \in \{1,\ldots,10\}$, a holistic overall score, and a boolean goal-achieved flag. The overall score is LLM-chosen rather than a deterministic function of the eight criterion scores; this is intentional to avoid over-weighting non-blocking issues.


\section{Reasoning-Trace Capture and Export}
\label{sec:reasoning}

Modern open-source models emit reasoning content either natively (via \texttt{reasoning\_content} in OpenAI-compatible vLLM responses) or inside explicit tags (\texttt{<think>}, \texttt{<reasoning>}). \sys{} captures all three modes via a reasoning extractor module and preserves them in the exported JSONL.

\subsection{Distillation Use Case}

The primary downstream application is reasoning distillation: generate $N$ scored conversations where the agent under test is a large reasoning model, filter to the top decile by judge score, and fine-tune a smaller student model on the filtered set. The student acquires teacher-quality reasoning behavior at a fraction of the deployment cost. \sys{} makes this operational by: (i) capturing reasoning traces as a first-class field, not requiring regex extraction; (ii) exporting in OpenAI function-calling JSONL unchanged; (iii) carrying per-sample judge scores in metadata for filtering.

\subsection{Export Format}

One conversation is one line; one line is one \texttt{\{messages, tools, metadata\}} object. Three properties: reasoning content is a first-class field on every assistant message; terminal actions do not emit spurious tool-response messages; metadata carries judge scores, scenario ID, persona profile, and complexity tier for downstream filtering.


\section{Scenario Validation Pipeline}
\label{sec:validation}

LLM-authored scenarios are fast but unreliable. A scenario might declare that the user wants to dispute a charge while the system facts contain no matching transaction. Running generation with such a scenario produces plausible-looking conversations that teach hallucination.

The scenario validator is a two-stage pipeline:

\paragraph{Stage 1: Programmatic checks.} JSON-Schema validation, referential integrity between user facts and system facts, intent-specific invariants (\eg{}, a dispute intent must reference at least one transaction), and near-duplicate detection via gestalt matching:
\begin{equation}
    \text{sim}(x, y) = 2M / (|x| + |y|)
\end{equation}
where $M$ is the total length of matching subsequences. Scenarios with $\text{sim} \geq 0.85$ (descriptions) or $\geq 0.90$ (goals) are flagged.

\paragraph{Stage 2: LLM semantic checks.} A frontier model answers six questions about the scenario: goal achievability, amount/ID consistency, final-state plausibility, edge-case validity, state exposure, and persona-intent fit. Consensus voting across $N$ models uses a pass ratio threshold of $\geq 0.5$. Issues are graded critical/error/warning/info; only scenarios with zero critical and zero error findings are promoted into a generation corpus.

\FloatBarrier

\section{Experiments and Results}
\label{sec:results}

We report results on three corpora produced by \sys{}, totaling 64,698 evaluated conversations.

\subsection{Corpora}

Before presenting results, we define the three evaluation corpora. The CRM (Customer Relationship Management) AI Assistant corpus refers to a domain-specific conversational agent dataset designed to evaluate tool-augmented assistant performance across customer-facing scenarios.

\begin{table}[h!]
\centering
\small
\setlength{\tabcolsep}{3pt}
\begin{tabular}{@{}lccl@{}}
\toprule
\textbf{Corpus} & \textbf{$n$} & \textbf{Scen.} & \textbf{Purpose} \\
\midrule
Mixed (Mar 2026)      & 49,331 & 312 & Training pool \\
CRM Train (Apr 2026)  & 12,224 & 77  & Domain fine-tune \\
CRM Golden (Apr 2026) &  3,143 & 20  & Benchmark (0 overlap) \\
\bottomrule
\end{tabular}
\caption{Corpus inventory. The CRM Golden Set has zero scenario overlap with CRM Training Set, enabling clean train/test measurement.}
\label{tab:corpora}
\end{table}

\subsection{Overall Score Distribution}

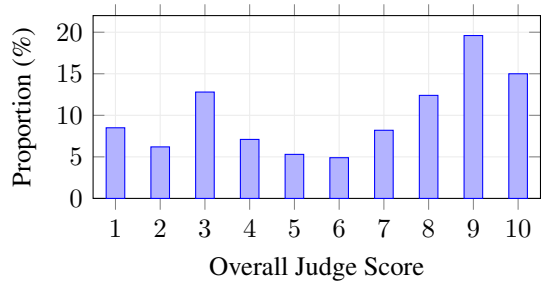
\begin{figure}[t]
\centering
\begin{tikzpicture}
\begin{axis}[
    ybar,
    width=7.5cm, height=4cm,
    xlabel={Overall Judge Score},
    ylabel={Proportion (\%)},
    xlabel style={font=\small},
    ylabel style={font=\small},
    xmin=0.5, xmax=10.5,
    ymin=0, ymax=22,
    xtick={1,2,3,4,5,6,7,8,9,10},
    ytick={0,5,10,15,20},
    bar width=7pt,
    grid=major,
    grid style={gray!15},
    every axis plot/.append style={fill=stateblue!65, draw=stateblue},
    tick label style={font=\small},
]
\addplot coordinates {
    (1, 8.5) (2, 6.2) (3, 12.8) (4, 7.1) (5, 5.3) (6, 4.9) (7, 8.2) (8, 12.4) (9, 19.6) (10, 15.0)
};
\end{axis}
\end{tikzpicture}
\caption{Overall judge score distribution across 49,331 conversations. The bimodal shape reflects binary agentic success. Mean 6.54, median 7.0, $\sigma$=2.25.}
\label{fig:score_dist}
\end{figure}

Across the 49,331-sample mixed corpus, mean overall score is 6.54 (median 7.0, $\sigma$=2.25). The distribution is bimodal with peaks near 3 and 9 (\Cref{fig:score_dist}), reflecting a two-population pattern: conversations where the agent either cleanly solves the scenario or fails at an identifiable step. The middle bucket is relatively empty as agentic success tends to be binary.

\subsection{Per-Criterion Analysis}

\begin{table}[h!]
\centering
\small
\begin{tabular}{@{}lcccc@{}}
\toprule
\textbf{Criterion} & \multicolumn{2}{c}{\textbf{Mixed (49K)}} & \multicolumn{2}{c}{\textbf{CRM Train}} \\
\cmidrule(lr){2-3} \cmidrule(lr){4-5}
 & $\mu$ & $\sigma$ & $\mu$ & $\sigma$ \\
\midrule
Goal achievement  & 5.52 & 3.24 & 5.26 & 2.03 \\
Tool usage        & 6.79 & 2.81 & 5.13 & 1.77 \\
Tool-call halluc. & \textbf{9.66} & 1.15 & 8.62 & 1.99 \\
Reasoning quality & 6.12 & 2.35 & 5.95 & 1.70 \\
Reasoning halluc. & \textbf{9.11} & 1.70 & 7.59 & 2.23 \\
Communication     & 6.37 & 2.18 & 7.61 & 1.14 \\
Consistency       & 8.47 & 1.68 & ---$^\dagger$ & --- \\
Error handling    & 6.13 & 2.46 & ---$^\dagger$ & --- \\
\bottomrule
\end{tabular}
\caption{Per-criterion means. Hallucination axes are near-ceiling, validating the state-grounding invariant. $^\dagger$Replaced by domain-specific axes in the CRM project.}
\label{tab:criterion_means}
\end{table}

The two hallucination axes sit at 9.66 and 9.11 in the mixed corpus and these are the axes the state manager is built to protect. Goal achievement at 5.52 is the residual hard problem: not hallucination, but reasoning-under-uncertainty and persona handling.

\subsection{Persona Variation}

The persona model produces measurable behavioral variation: across eight persona profiles with $\geq$400 samples each, goal-achievement rates range from 51.2\% to 67.0\%, a 15.8 percentage-point spread. This confirms persona is a real lever on agent performance, not decoration.

\subsection{Correlation Structure}

\begin{figure}[t]
\centering
\begin{tikzpicture}
\small
\def\sz{0.70}
\foreach \i/\name in {0/GA,1/TU,2/TH,3/RQ,4/RH,5/CQ,6/CO,7/EH}{
    \node[font=\scriptsize\bfseries] at (\i*\sz, 8.2*\sz) {\name};
    \node[font=\scriptsize\bfseries] at (-1.0*\sz, {(7-\i)*\sz}) {\name};
}
\foreach \i/\j/\val/\col in {
    0/0/1.00/100, 0/1/0.72/72, 0/2/0.21/21, 0/3/0.83/83, 0/4/0.21/21, 0/5/0.86/86, 0/6/0.68/68, 0/7/0.85/85,
    1/1/1.00/100, 1/2/0.35/35, 1/3/0.71/71, 1/4/0.30/30, 1/5/0.70/70, 1/6/0.58/58, 1/7/0.69/69,
    2/2/1.00/100, 2/3/0.19/19, 2/4/0.66/66, 2/5/0.18/18, 2/6/0.25/25, 2/7/0.20/20,
    3/3/1.00/100, 3/4/0.23/23, 3/5/0.88/88, 3/6/0.72/72, 3/7/0.87/87,
    4/4/1.00/100, 4/5/0.20/20, 4/6/0.28/28, 4/7/0.22/22,
    5/5/1.00/100, 5/6/0.74/74, 5/7/0.89/89,
    6/6/1.00/100, 6/7/0.71/71,
    7/7/1.00/100
}{
    \fill[stateblue!\col] (\i*\sz-\sz/2, {(7-\j)*\sz-\sz/2}) rectangle ++(\sz,\sz);
    \fill[stateblue!\col] (\j*\sz-\sz/2, {(7-\i)*\sz-\sz/2}) rectangle ++(\sz,\sz);
    \pgfmathparse{\col > 40 ? "white" : "stateblue"}
    \edef\txtcol{\pgfmathresult}
    \node[font=\tiny, text=\txtcol] at (\i*\sz, {(7-\j)*\sz}) {\val};
    \ifnum\i=\j\else
        \node[font=\tiny, text=\txtcol] at (\j*\sz, {(7-\i)*\sz}) {\val};
    \fi
}
\end{tikzpicture}
\caption{Pearson correlation across 8 criteria (49K samples). TH and RH form their own cluster ($\rho=0.66$) but correlate weakly with GA ($\rho \approx 0.21$). GA=Goal Achievement, TU=Tool Usage, TH=Tool-call Hallucination, RQ=Reasoning Quality, RH=Reasoning Hallucination, CQ=Communication Quality, CO=Consistency, EH=Error Handling.}
\label{fig:correlation}
\end{figure}

Two observations from the correlation matrix (\Cref{fig:correlation}): (i) reasoning quality, communication quality, and error handling form a tight cluster ($\rho > 0.85$); (ii) the hallucination axes correlate with each other ($\rho = 0.66$) but only weakly with goal achievement ($\rho \approx 0.21$). This confirms hallucination is its own failure mode, not a symptom of poor reasoning.

\subsection{Train-vs-Golden Gap Analysis}

\begin{figure}[h!]
\centering
\begin{tikzpicture}
\begin{axis}[
    ybar,
    width=7.5cm, height=4cm,
    xlabel={Overall Judge Score},
    ylabel={Proportion (\%)},
    xlabel style={font=\small},
    ylabel style={font=\small},
    xmin=0.5, xmax=10.5,
    ymin=0, ymax=28,
    xtick={1,2,3,4,5,6,7,8,9,10},
    bar width=4pt,
    grid=major,
    grid style={gray!15},
    tick label style={font=\small},
    legend style={font=\scriptsize, at={(0.98,0.95)}, anchor=north east, draw=gray!40},
    legend cell align={left},
]
\addplot[fill=stateblue!50, draw=stateblue] coordinates {
    (1,8.5)(2,6.2)(3,12.8)(4,7.1)(5,5.3)(6,4.9)(7,8.2)(8,12.4)(9,19.6)(10,15.0)
};
\addplot[fill=agentgreen!50, draw=agentgreen] coordinates {
    (1,1.2)(2,4.5)(3,8.9)(4,12.1)(5,14.8)(6,16.2)(7,15.5)(8,12.8)(9,8.7)(10,5.3)
};
\addplot[fill=toolorange!50, draw=toolorange] coordinates {
    (1,3.1)(2,7.8)(3,14.2)(4,16.5)(5,17.1)(6,15.3)(7,11.2)(8,7.9)(9,4.6)(10,2.3)
};
\legend{Mixed ($\mu$=6.54), CRM train ($\mu$=6.25), CRM golden ($\mu$=5.44)}
\end{axis}
\end{tikzpicture}
\caption{Overall-score distributions for the three corpora. The mixed corpus is bimodal; CRM training is roughly normal; CRM golden is left-shifted (harder).}
\label{fig:three_corpora}
\end{figure}
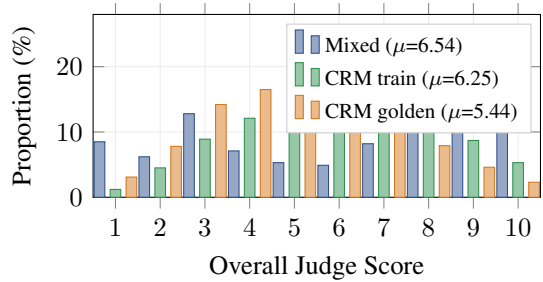

\begin{table}[h!]
\centering
\small
\begin{tabular}{@{}lccc@{}}
\toprule
\textbf{Criterion} & \textbf{Train} & \textbf{Golden} & \textbf{$\Delta$} \\
\midrule
Overall score      & 6.25 & 5.44 & $-$0.81 \\
Goal achievement   & 5.26 & 4.38 & $-$0.88 \\
Tool usage         & 5.13 & 4.29 & $-$0.84 \\
Tool-call halluc.  & 8.62 & 7.93 & $-$0.69 \\
Reasoning quality  & 5.95 & 5.06 & $-$0.89 \\
Reasoning halluc.  & 7.59 & 6.43 & $-$1.16 \\
Communication      & 7.61 & 7.20 & $-$0.41 \\
Booking compliance & 5.73 & 4.93 & $-$0.80 \\
Channel safety     & 9.37 & 9.01 & $-$0.36 \\
\bottomrule
\end{tabular}
\caption{Train-vs-golden per-criterion gap. Content criteria drop the most; safety criteria are robust across splits.}
\label{tab:train_golden}
\end{table}

The golden set scores lower on every criterion (\Cref{tab:train_golden}), with the largest drops on reasoning hallucination ($-$1.16), reasoning quality ($-$0.89), and booking-flow compliance ($-$0.80). Channel-safety compliance barely moves ($-$0.36) because safety rules are scenario-independent. A model that closes this gap is genuinely generalizing, not memorizing scenario patterns.

\subsection{Turn-Count and Scale}

Conversations finishing in 1--2 turns often hit the terminal action without enough context; those past 7--8 turns typically indicate the agent is stuck in a loop. The sweet spot is 4--6 turns, matching production traffic patterns.

\FloatBarrier

\section{Comparison with External Systems}
\label{sec:comparison}

We compare \sys{} against eight external systems across the four capabilities that define our platform. \Cref{tab:comparison} summarizes the landscape. Claims are verified against vendor documentation, peer-reviewed papers, or official repositories.

\begin{table}[h!]
\centering
\small
\setlength{\tabcolsep}{2.5pt}
\begin{tabular}{@{}lcccc@{}}
\toprule
\textbf{System} & \textbf{MT} & \textbf{SG} & \textbf{MA} & \textbf{Judge} \\
\midrule
NeMo Designer       & \checkmark & Partial$^*$ & ---     & Plugin \\
Gretel              & Tabular    & ---         & ---     & --- \\
MOSTLY AI           & Tabular    & ---         & ---     & --- \\
OpenAI Evals        & \checkmark & User-coded  & ---     & \checkmark \\
AutoGen \cite{wu2023autogen} & Runtime & Implicit & Runtime & --- \\
LangGraph           & Runtime    & Implicit    & Runtime & --- \\
$\tau$-bench \cite{yao2024taubench} & \checkmark & Scripted & --- & \checkmark \\
Matrix \cite{mei2024matrix} & \checkmark & Partial & Research & --- \\
\midrule
\textbf{\sys{} (ours)} & \textbf{\checkmark} & \textbf{\checkmark} & \textbf{\checkmark} & \textbf{\checkmark} \\
\bottomrule
\end{tabular}
\caption{Capability comparison. MT=Multi-turn gen., SG=State-grounded tool sim., MA=Multi-agent data gen. $^*$Prompt-templated only. ``Runtime''=inference-time only. ``---''=not documented.}
\label{tab:comparison}
\end{table}

\paragraph{Honest positioning.}
\sys{} is not the right tool for every synthetic data task. Gretel and MOSTLY AI have more mature differential-privacy guarantees for tabular data. NeMo Data Designer achieves higher tokens/second for single-turn instruction generation. $\tau$-bench has more rigorous task design for academic benchmarking. The defensible claim is narrower: for agentic training data that requires multi-turn generation, state-grounded tool simulation, hierarchical multi-agent support, and built-in judge scoring, we could not find a single publicly available system that ships all four.

\FloatBarrier

\section{Limitations and Future Work}
\label{sec:limitations}

\paragraph{Judge calibration.}
The 8-axis judge has not been calibrated against human gold-standard annotations. While inter-axis correlation structure (\Cref{fig:correlation}) is informative, absolute score levels may carry systematic bias from the judge model. Collecting 1--2K human-rated samples and learning per-criterion bias corrections is planned.

\paragraph{Persona covariance.}
Trait correlations are currently rule-described, not learned. Fitting a covariance matrix from the 64K sample corpus would yield more realistic trait co-occurrence without manual rule authoring.

\paragraph{Overall score transparency.}
The overall score is LLM-chosen rather than a deterministic function of the eight criterion scores. This creates a transparency concern for downstream consumers. Future work includes either making the aggregation explicit or documenting the empirical relationship.

\paragraph{State manager fidelity.}
The state manager runs at temperature 0 but is still an LLM, not a database. Complex state updates (multi-record transactions, partial rollbacks) may introduce subtle inconsistencies. Hybrid approaches with deterministic state logic for structured operations are under investigation.

\paragraph{Near-term extensions.}
Tool-failure injection (timeouts, rate limits, malformed responses) to train graceful degradation. Preference-pair generation for RLHF/DPO from same-scenario winning/losing conversations. Multi-teacher ensembling across frontier models per scenario.


\section{Conclusion}
\label{sec:conclusion}

We presented \sys{}, a synthetic data generation platform for tool-augmented LLM agents. The core contribution is an authoritative state manager that enforces a backend-is-truth invariant across the generation loop, achieving 9.66/10 on tool-call hallucination across 49K+ samples. The platform extends to hierarchical multi-agent generation through a sub-agents-as-tools pattern with shared state, supports persona-conditioned user simulation via a 23-dimensional trait vector with measurable downstream impact (15.8 percentage-point spread on goal achievement), and produces per-sample 8-axis judge scores that enable fine-grained corpus filtering.

Across 64,698 evaluated conversations spanning 3 production corpora, the train-vs-golden gap analysis confirms the output is not memorization bait: content criteria drop on held-out scenarios while safety criteria remain robust. Comparison with eight external systems identifies the capability combination of multi-turn generation, state-grounded tool simulation, multi-agent data generation, and built-in multi-axis judging as the architectural contribution that no single existing platform provides.

The platform reduces the timeline for training-data generation from months of human annotation to days of configured generation, making it practical for organizations deploying multiple agentic products in parallel.


\bibliographystyle{plain}

\end{document}